\documentclass[10pt,twocolumn,letterpaper]{article}

\usepackage{cvpr}              

\usepackage{makecell}
\usepackage{rotating}
\usepackage{colortbl}
\usepackage{threeparttable}
\usepackage{footmisc}
\usepackage{multirow}

\newcommand{\Name}{{OccMamba}\xspace}
\newcommand\blfootnote[1]{%
  \begingroup
  \renewcommand\thefootnote{}\footnote{#1}%
  \addtocounter{footnote}{-1}%
  \endgroup
}

\definecolor{cvprblue}{rgb}{0.21,0.49,0.74}
\usepackage[pagebackref,breaklinks,colorlinks,allcolors=cvprblue]{hyperref}

\def\paperID{4501} 
\def\confName{CVPR}
\def\confYear{2025}

\title{OccMamba: Semantic Occupancy Prediction with State Space Models}

\author{
    Heng Li\textsuperscript{\rm 1},
    Yuenan Hou\textsuperscript{\rm 2}*,
    Xiaohan Xing\textsuperscript{\rm 3},
    Yuexin Ma\textsuperscript{\rm 4},
    Xiao Sun\textsuperscript{\rm 2},
    Yanyong Zhang\textsuperscript{\rm 1 5}*\\
    \textsuperscript{\rm 1} University of Science and Technology of China \\
    \textsuperscript{\rm 2} Shanghai AI Laboratory  
    \textsuperscript{\rm 3} Stanford University 
    \textsuperscript{\rm 4} ShanghaiTech University \\
    \textsuperscript{\rm 5} Institute of Artificial Intelligence, Hefei Comprehensive National Science Center \\
    {\tt\small 
        li\_heng@mail.ustc.edu.cn,
        houyuenan@pjlab.org.cn,
        xhxing@stanford.edu
    }\\
    {\tt\small 
        mayuexin@shanghaitech.edu.cn, 
        sunxiao@pjlab.org.cn, 
        yanyongz@ustc.edu.cn
    }
}

\begin{document}
\maketitle
\begin{abstract}

\vspace{-5mm}

\blfootnote{This work was supported by the National Natural Science Foundation of China (No. 62332016) and the Key Research Program of Frontier Sciences, CAS (No. ZDBS-LY-JSC001).}
\blfootnote{* The corresponding authors.}

Training deep learning models for semantic occupancy prediction is challenging due to factors such as a large number of occupancy cells, severe occlusion, limited visual cues, complicated driving scenarios, etc. Recent methods often adopt transformer-based architectures given their strong capability in learning input-conditioned weights and long-range relationships. However, transformer-based networks are notorious for their quadratic computation complexity, seriously undermining their efficacy and deployment in semantic occupancy prediction. Inspired by the global modeling and linear computation complexity of the Mamba architecture, we present the first Mamba-based network for semantic occupancy prediction, termed \Name. Specifically, we first design the hierarchical Mamba module and local context processor to better aggregate global and local contextual information, respectively. Besides, to relieve the inherent domain gap between the linguistic and 3D domains, we present a simple yet effective 3D-to-1D reordering scheme, i.e., height-prioritized 2D Hilbert expansion. It can maximally retain the spatial structure of 3D voxels as well as facilitate the processing of Mamba blocks.
Endowed with the aforementioned designs, our \Name is capable of directly and efficiently processing large volumes of dense scene grids, achieving state-of-the-art performance across three prevalent occupancy prediction benchmarks, including OpenOccupancy, SemanticKITTI, and SemanticPOSS. Notably, on OpenOccupancy, our \Name outperforms the previous state-of-the-art Co-Occ by \textbf{5.1\%} IoU and \textbf{4.3\%} mIoU, respectively. 
Our implementation is open-sourced and available at: \href{https://github.com/USTCLH/OccMamba}{https://github.com/USTCLH/OccMamba}.

\end{abstract}    
\section{Introduction}
\label{sec:introduction}



\begin{figure*}[tp]
    \centering
    \begin{tabular}{ccc}
        \includegraphics[width=0.3\textwidth]{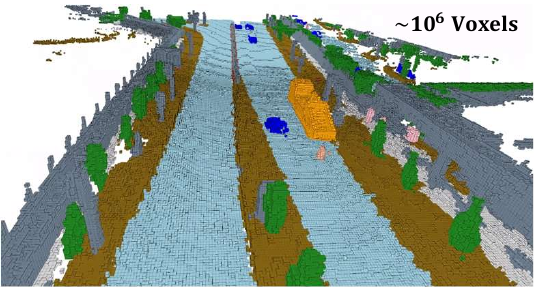} &
        \includegraphics[width=0.33\textwidth]{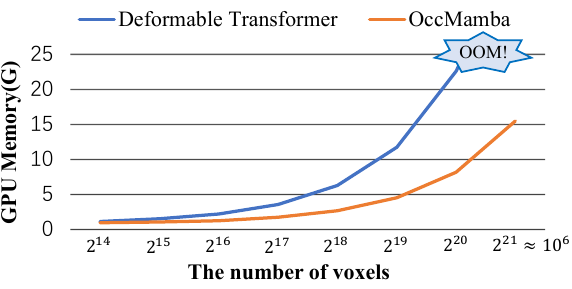} &
        \includegraphics[width=0.3\textwidth]{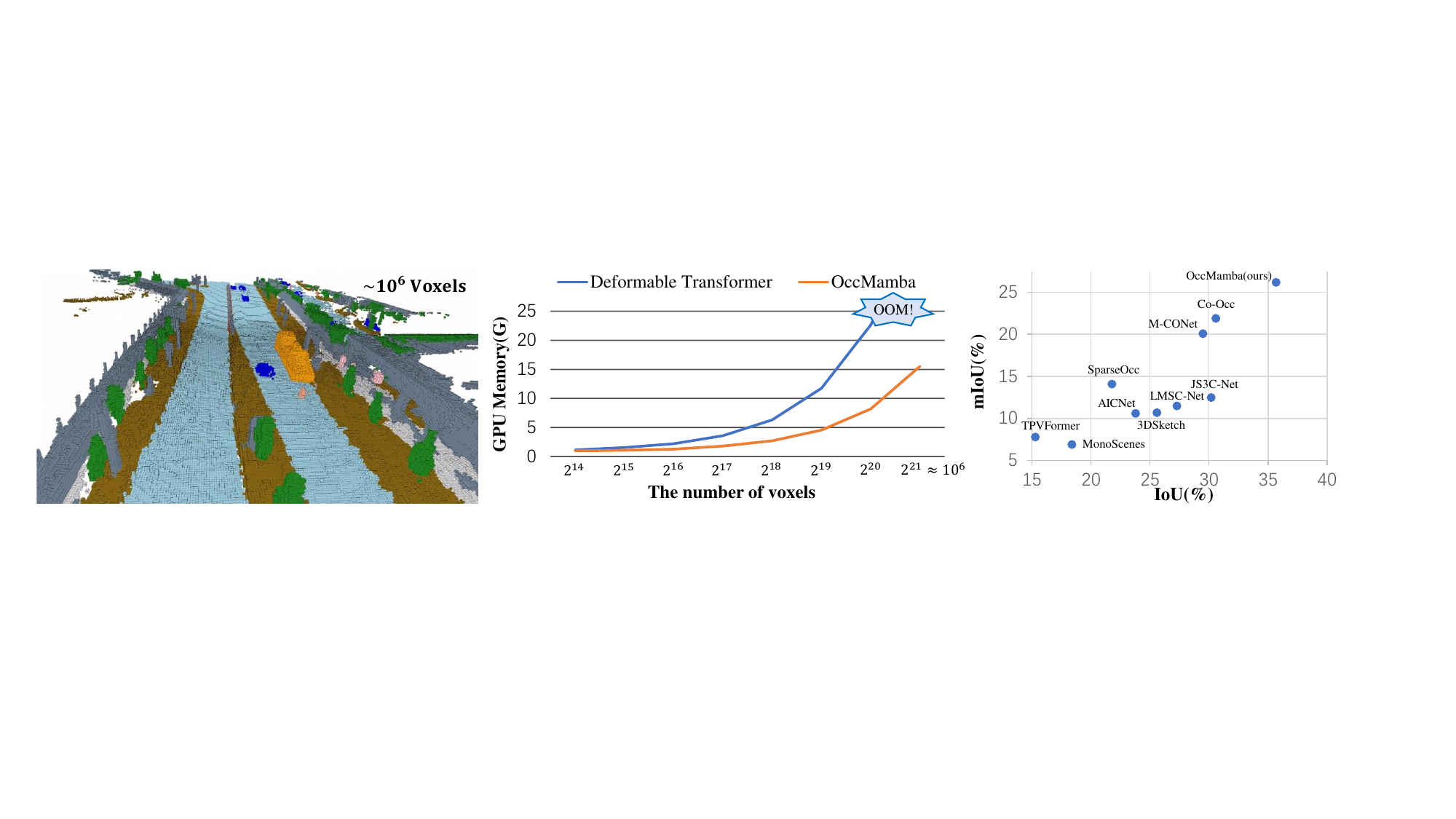} \\
        \parbox[c]{0.3\textwidth}{\centering (a) A large number of voxel grids} &
        \parbox[c]{0.33\textwidth}{\centering (b) GPU memory consumption} &
        \parbox[c]{0.3\textwidth}{\centering (c) Performance comparison} \\
    \end{tabular}
    \caption{(a) Challenges in semantic occupancy prediction, (b) comparison in GPU memory consumption, and (c) performance comparison on OpenOccupancy validation set. Our \Name demonstrates high efficiency in handling a large number of voxel grids, outperforming all other semantic occupancy predictors, such as Co-Occ~\cite{co-occ} and M-CONet~\cite{openoccupancy}.}
    \label{fig:introduction}
    \vspace{-5mm}
\end{figure*}

Semantic occupancy prediction is becoming an indispensable component in autonomous driving, augmented reality, robotics, etc., which estimates the occupancy and categorical labels of the surrounding environment~\cite{ssc_survey}. It faces many challenges, such as the enormous number of occupancy grids, severe occlusion, limited visual clues as well as complex driving scenarios~\cite{vision_ssc_survey}.

Recent attempts, such as MonoScene~\cite{monoscene} and JS3C-Net~\cite{js3cnet}, have made progress in addressing these challenges, but limitations remain due to their reliance on uni-modal inputs. Multi-modal approaches, such as FusionOcc~\cite{fusionocc} and M-CONet~\cite{openoccupancy}, offer improvements, yet they struggle to capture global information due to the inherent deficiency of CNN architectures. Despite the success of transformer-based models~\cite{voxformer, occnet, occformer}, they suffer from high computational complexity, particularly when processing a large number of voxel grids.

Mamba~\cite{mamba}, which is an important variant of state space models, emerges as a promising next-generation structure for replacing the transformer architecture. In semantic occupancy prediction, we anticipate leveraging its global modeling capabilities to better handle the complex scenarios and overcome limited visual cues, while its linear computation complexity helps manage the enormous number of occupancy grids efficiently.
However, it is primarily designed for language modeling, 
with 1D input data, whereas the input data for semantic occupancy prediction is 3D voxels. The use of dense voxels in large scenes makes the deployment of Mamba even more challenging. When converting 3D data into a 1D format, the inherent spatial relationships between adjacent voxels in the 3D space are missing, causing neighboring voxels to become far apart in the 1D sequence. 
This spatial separation undermines Mamba’s ability to effectively understand local and global scene contexts, hampering the prediction performance. To address this issue, developing an effective reordering strategy is crucial to preserving spatial proximity in the transformation from 3D to 1D. Several attempts have been made to achieve this, for instance, Point Mamba~\cite{pointmamba} rearranges the point clouds according to the 3D Hilbert curve~\cite{hilbert} and concatenates the features of the reordered points. While these advancements have shown potential in the point cloud classification and segmentation field, the exploration of Mamba-based architectures is still in its infancy in outdoor semantic occupancy prediction tasks.

In this work, we present the first Mamba-based network for semantic occupancy prediction, which we refer to as \Name. Owing to the global modeling and linear computation complexity of Mamba, our \Name can efficiently process a large number of voxel grids given limited computation resources, as shown in Fig.~\ref{fig:introduction}(a)-(b). To effectively utilize the global and local information hidden in the input voxel grids, we design the hierarchical Mamba module and local context processor. To facilitate the processing of Mamba blocks as well as preserve the spatial structure of the 3D input data, we present a simple yet effective 3D-to-1D reordering policy, i.e., height-prioritized 2D Hilbert expansion. The designed policy sufficiently utilizes the categorical clues in the height information as well as the spatial prior in the XY plane. In this way, \Name effectively exploits LiDAR and camera cues, enabling effective fusion and processing of voxel features extracted from these sources without compression. This uncompressed voxel feature processing method, combined with Mamba's global modeling capabilities, enhances \Name with the occlusion reasoning capability, particularly in complex driving scenarios. 
We perform extensive experiments on three semantic occupancy prediction benchmarks, i.e., OpenOccupancy~\cite{openoccupancy}, SemanticKITTI~\cite{semantickitti} and SemanticPOSS~\cite{semanticposs}. Our \Name consistently outperforms state-of-the-art algorithms in all benchmarks. It is noteworthy that on OpenOccupancy, partly shown in Fig.~\ref{fig:introduction}(c), our \Name surpasses the previous state-of-the-art Co-Occ~\cite{co-occ} by \textbf{5.1\%} IoU and \textbf{4.3\%} mIoU, respectively.

The contributions are summarized as follows:
\begin{itemize}[leftmargin=*]
\item To our knowledge, we design the first Mamba-based network for outdoor semantic occupancy prediction. It possesses global modeling capability with linear computation complexity, which is crucial to the processing of a large number of voxel grids. We design the hierarchical Mamba module and local context processor to better aggregate global and local contextual information, respectively. 

\item To facilitate the processing of Mamba blocks as well as maximally retain the original spatial structure of 3d voxels, we design a simple yet effective reordering policy that projects the point clouds into 1D sequences.   

\item Our \Name achieves the best performance on three popular semantic occupancy prediction benchmarks.

\end{itemize}
\section{Related work}
\label{sec:relatedwork}

\textbf{Semantic occupancy prediction.} The objective of semantic occupancy prediction is to estimate the occupancy and semantic labels of 3D spaces using various types of input signals~\cite{ssc_survey,survey_3d}. Given that LiDAR and RGB-D camera can provide accurate spatial measurement, many studies relied on such detailed geometric data like LiDAR points~\cite{local-DIFs, scpnet, lmscnet, js3cnet} and RGB-D images~\cite{3dsketch, aicnet, two_stream, AMFNet}. Meanwhile, image-based methods, increasingly popular due to camera accessibility, such as MonoScene~\cite{monoscene} and TPVFormer~\cite{tpvformer}, estimate occupancy of the environment using RGB images. However, inaccurate depth estimation in RGB images results in lower performance than LiDAR-based models. In response, multi-modal fusion combining both modalities has attracted significant attention. CONet~\cite{openoccupancy} and Co-Occ~\cite{co-occ} demonstrate the advantages of combining modalities, enhancing precision and reliability. 
Furthermore, recent methods have shifted from CNNs to integrating transformers, as seen in OccFormer's dual-path transformer~\cite{occformer} and OccNet's cascade refinement~\cite{occnet} reduce the transformer's computational demands. 






\noindent \textbf{Multi-modal fusion.} Multi-modal fusion aims to make the strength of different input modalities, thus making more accurate and robust perception~\cite{mm_survey,aicnet,bevfusion,unitr,logonet,uniseg,hmfi}. For instance, AICNet~\cite{aicnet} integrates RGB and depth data using anisotropic convolutional networks to enhance the accuracy and completeness of semantic occupancy prediction. PointPainting~\cite{pointpainting} performs semantic segmentation on RGB images and attaches semantic probabilities to LiDAR points, thereby enriching the point cloud with rich semantic information. For semantic occupancy prediction, images and point clouds are two prevalent input signals. Recent trends in the occupancy prediction field favour the multi-modal fusion as it utilizes the strength of both signals.


\begin{figure*}[!t]
    \centering
    \includegraphics[width=\linewidth]{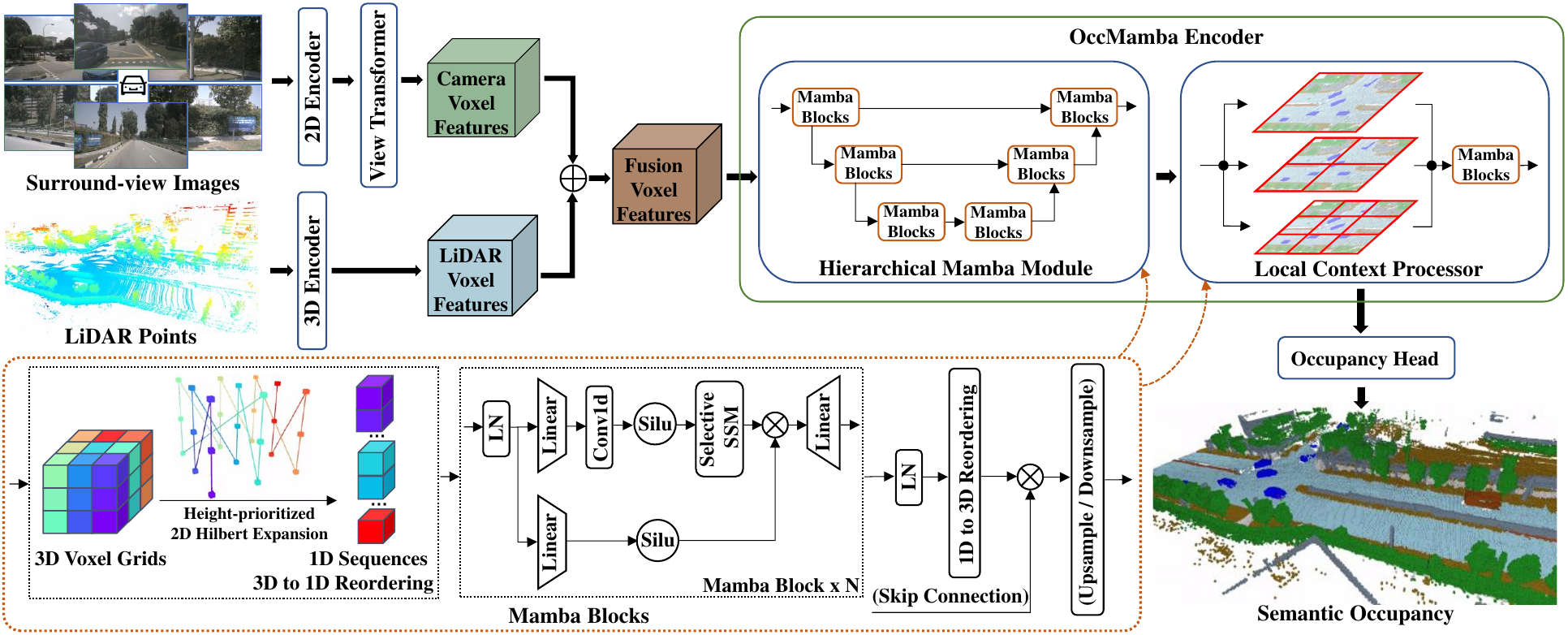}
    \caption{Schematic overview of our \Name. Given surround-view images and LiDAR point clouds, 
    we first employ the 2D encoder and 3D encoder to process them, obtaining camera features and LiDAR voxel features, respectively. View transformer is utilized to project camera features to camera voxel features. The camera and LiDAR voxel features are then fused and sent to the hierarchical Mamba module, where the proposed height-prioritized 2D Hilbert expansion reordering is used to maximally utilize the spatial clues of voxels. Besides, local context processor is designed to divide the Mamba features into multiple windows along the XY plane, further enhancing the local semantic information. Eventually, the Mamba features are fed to the occupancy head, producing semantic occupancy predictions.} 
    \label{fig:pipeline}
    \vspace{-5mm}
\end{figure*}

\noindent \textbf{State space models.} The transformer~\cite{transformer} has reshaped the computer vision field but suffers from the quadratic computation complexity~\cite{transformer_survey}. To relieve this, more efficient operators like linear attention~\cite{linformer}, flash attention~\cite{flashattn} have been proposed. State Space Models (SSMs) such as Mamba~\cite{mamba}, S4~\cite{S4}, and S4nd~\cite{S4nd} are gaining prominence, with Mamba being notable for integrating selective mechanisms, which can effectively capture long-range dependencies and process large-scale data in linear time. This innovation has extended into computer vision domain through variants like VMamba~\cite{vmamba}, which includes a cross-scan module, and Vision Mamba~\cite{visionmamba}, which utilizes a bidirectional SSM. In point cloud processing, PointMamba~\cite{pointmamba} improves the global modeling of point clouds by rearranging input patches based on 3D Hilbert curve, while Point Mamba~\cite{point_mamba} employs an octree-based ordering for efficient spatial relationship capture. They demonstrate Mamba's proficiency in processing large scale 3D data. Building on these advancements, we attempt to design a Mamba-based network tailored for semantic occupancy prediction.
\section{Methodology}
\label{sec:methodology}


To efficiently and effectively process a large number of voxel grids in semantic occupancy prediction, we propose to take Mamba~\cite{mamba} as the basic building block that enjoys the benefit of both global modeling and linear computation complexity. To facilitate the processing of Mamba blocks, we design a simple yet effective reordering scheme, dubbed height-prioritized 2D Hilbert expansion, along with OccMamba Encoder. In the following sections, we first provide a brief review of state space models and Mamba in Sec.~\ref{subsec:pre}. Then, we present the framework overview of \Name in Sec.~\ref{subsec:network}. Thereafter, we provide a detailed explanation of the reordering scheme, the hierarchical Mamba module and local context processor in Sec.~\ref{subsec:mamba}. Eventually, the training objective is presented in Sec.~\ref{subsec:obj}.

\subsection{Preliminaries}
\label{subsec:pre}

\noindent \textbf{State space models.} Before introducing the Mamba module, we first have a brief review of the state space models (SSMs). SSMs are inspired by the control theory and maps the system input to the system output through the hidden state. This approach allows for effectively handling sequences of information. When the input is discrete, the mathematical representation of SSMs is given as follows:

\vspace{-5mm}
\begin{align}
h_{k} &= \overline{A} h_{k-1} + \overline{B} x_{k}, \\
y_{k} &= \overline{C} h_{k},
\end{align}

\noindent where $k$ is the sequence number, $\overline{A}$, $\overline{B}$, $\overline{C}$ are matrices that represent the discretized parameters of the model, which involve the sampling step $\Delta$. The $x_k$, $y_k$, and $h_k$ denote the input, output, and hidden state of the system, respectively. As an improvement, Structured State Space Sequence Models (S4)~\cite{S4} optimize traditional SSMs by introducing structured matrices, which allow the system dynamics, involving matrices $\overline{A}$, $\overline{B}$, and $\overline{C}$, to be parameterized in a way that significantly improves computational efficiency and scalability for large sequences. By leveraging these structured matrices, S4 achieves reduced computational complexity without sacrificing accuracy.

\noindent \textbf{Mamba Module.} Mamba introduces an adaptation to the S4 models~\cite{S4}, where it makes the matrices $\overline{B}$ and $\overline{C}$, as well as the sampling size $\Delta$, dependent on the input. This dependency arises from incorporating the sequence length and batch size of the input, enabling dynamic adjustments of these matrices for each input token. By this way, Mamba allows $\overline{B}$ and $\overline{C}$ to dynamically influence the state transition conditioned on the input, enhancing the model's content-awareness. Additionally, Mamba incorporates optimizations for the scan operation and a hardware-aware algorithm, enabling efficient parallel computation.


\subsection{Framework overview}
\label{subsec:network}

Figure~\ref{fig:pipeline} depicts the pipeline of our \Name, which is built upon M-CONet~\cite{openoccupancy}. 

\noindent \textbf{Multi-modal visual encoders.}  Taking both point cloud and multi-view images as input, \Name processes each modality with respective visual encoder. Specifically, for the LiDAR branch, we first voxelize the input point clouds $\mathbf{P}$~\cite{voxelnet} and then employ sparse-convolution-based LiDAR encoder $\mathbf{E}_\mathcal{L}$ ~\cite{second} to generate LiDAR voxel features $\mathbf{V}_\mathcal{L} \in \mathbb{R}^{B \times W_\mathcal{L} \times H_\mathcal{L} \times D_\mathcal{L} \times C_\mathcal{L}}$. For the image branch, we feed the multi-view images $\mathbf{I}$ to ResNet-based image encoder~\cite{resnet} which utilizes FPN~\cite{fpn} to aggregate multi-scale features, and then utilize the 2D-to-3D view transformer~\cite{bevfusion} to produce image voxel features $\mathbf{V}_\mathcal{C} \in \mathbb{R}^{B \times W_\mathcal{C} \times H_\mathcal{C} \times D_\mathcal{C} \times C_\mathcal{C}}$. Here, $B$ is the batch size, $W_\mathcal{L}$, $W_\mathcal{C}$, $H_\mathcal{L}$, $H_\mathcal{C}$, $D_\mathcal{L}$ and $D_\mathcal{C}$ are the spatial dimensions of the voxel features. $C_\mathcal{L}$ and $C_\mathcal{C}$ are the channel dimension of these voxel features. We ensure that $\mathbf{V}_\mathcal{L}$ and $\mathbf{V}_\mathcal{C}$ are identical in the spatial dimensions. Thereafter, these voxel features are concatenated along the channel dimensions:

\vspace{-3mm}
\begin{equation}
\mathbf{V}_\mathcal{F} = \text{concat}(\mathbf{V}_\mathcal{L}, \mathbf{V}_\mathcal{C}).
\end{equation}
\vspace{-5mm}


\noindent \textbf{\Name encoder.} To process features extracted from the multi-modal encoders, we designed the hierarchical Mamba module and local context processor. The former includes the Mamba encoder $\mathbf{E}_\mathcal{M}$ and Mamba decoder $\mathbf{D}_\mathcal{M}$, while the latter is denoted as $\mathbf{P}_\mathcal{L}$.
In the process, fusion voxel features $\mathbf{V}_\mathcal{F}$ are fed into them, which will be detailed in Sec.~\ref{subsec:mamba}. The transformation can be described by

\vspace{-5mm}
\begin{align}
\mathbf{V}_\mathcal{M} &= \mathbf{D}_\mathcal{M}(\mathbf{E}_\mathcal{M}(\mathbf{V}_\mathcal{F})), \\
\mathbf{V}_\mathcal{P} &= \mathbf{P}_\mathcal{L}(\mathbf{V}_\mathcal{M}),
\end{align}
\vspace{-5mm}

\noindent where $\mathbf{V}_\mathcal{M}$ and $\mathbf{V}_\mathcal{P}$ retains the same size as $\mathbf{V}_\mathcal{F}$. 

\noindent \textbf{Occupancy head.} We feed the processed features of \Name Encoder, i.e., $\mathbf{V}_\mathcal{P}$, to the coarse-to-fine module $\mathbf{F}_{c \rightarrow f}$~\cite{openoccupancy}, which interpolates the dimensions to match the size of the ground truth, and then employ a MLP to predict the category of each voxel grid. The process is presented in the following equation:

\vspace{-3mm}
\begin{equation}
\mathbf{O}_{\text{occ}} = \text{MLP}(\mathbf{F}_{c \rightarrow f}(\mathbf{V}_\mathcal{P})).
\end{equation}
\vspace{-2mm}

\begin{figure*}[tp]
    \centering
    \begin{tabular}{ccccc}
        \begin{turn}{90} Isometric view \end{turn} &
        \includegraphics[width=0.18\textwidth]{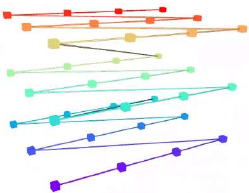} &
        \includegraphics[width=0.18\textwidth]{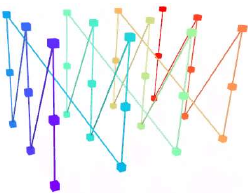} &
        \includegraphics[width=0.18\textwidth]{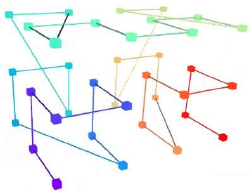} &
        \includegraphics[width=0.18\textwidth]{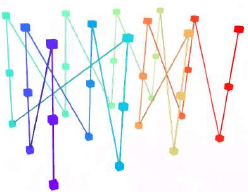} \\
        
        \begin{turn}{90} \quad Top view \end{turn} &
        \includegraphics[width=0.18\textwidth]{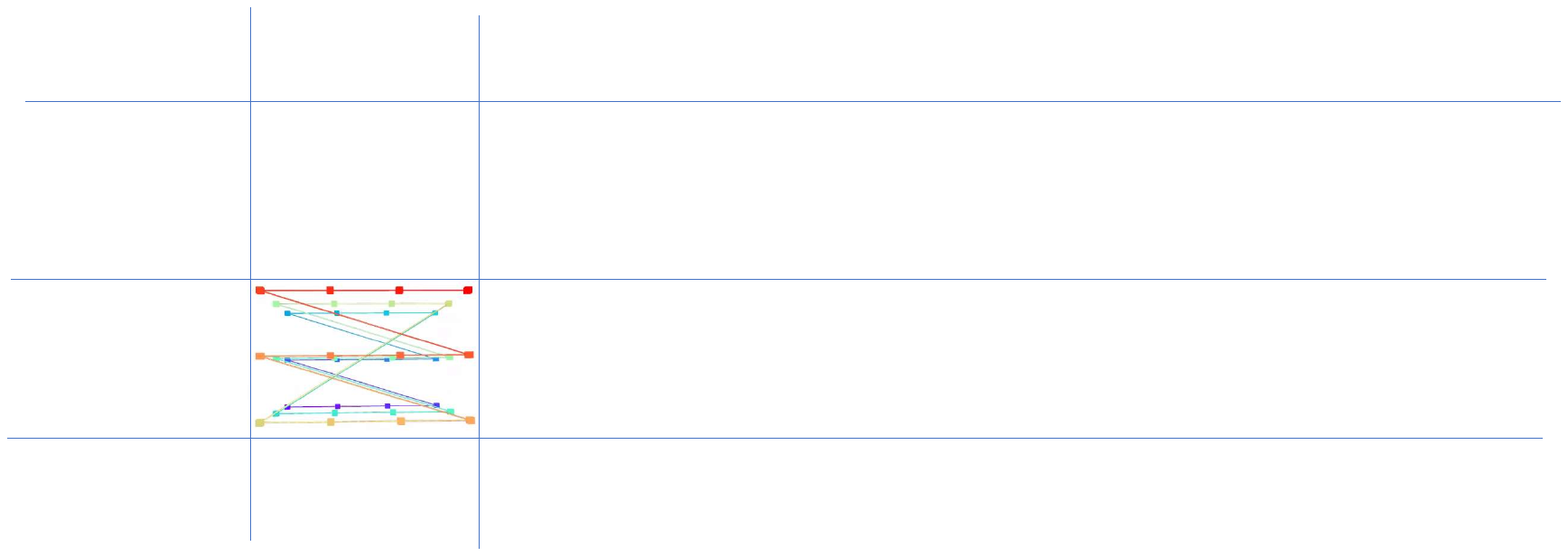} &
        \includegraphics[width=0.18\textwidth]{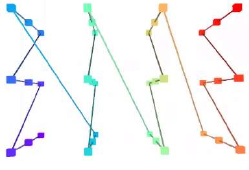} &
        \includegraphics[width=0.18\textwidth]{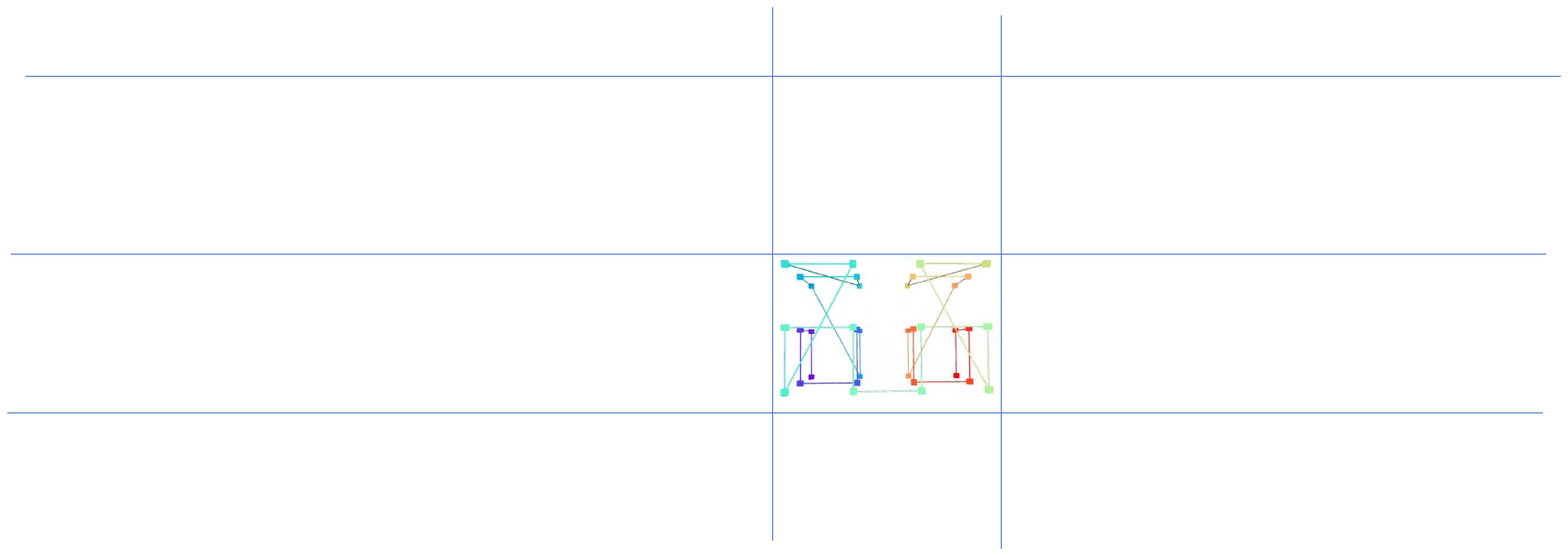} &
        \includegraphics[width=0.18\textwidth]{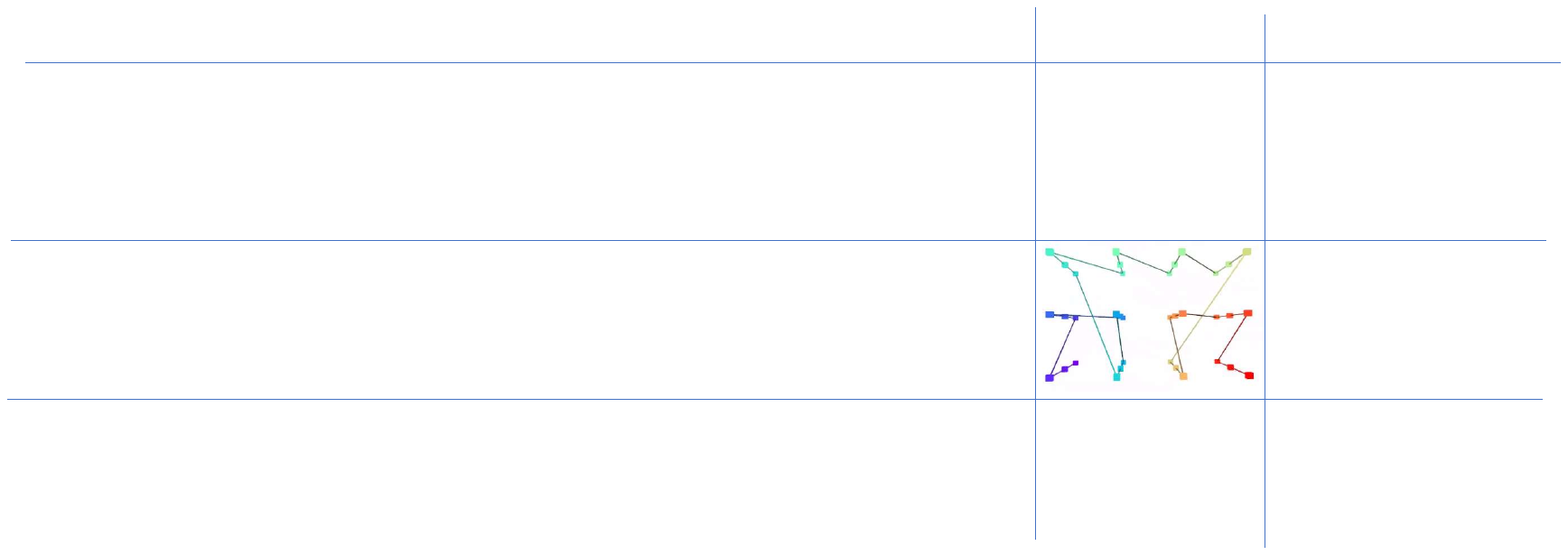} \\

         &
        \parbox[c]{0.18\textwidth}{\centering (a) XYZ sequence} &
        \parbox[c]{0.18\textwidth}{\centering (b) ZXY sequence} &
        \parbox[c]{0.18\textwidth}{\centering (c) 3D Hilbert curve} &
        \parbox[c]{0.18\textwidth}{\centering (d) Height-prioritized 2D Hilbert expansion} \\
    \end{tabular}
    \caption{Comparison between different reordering schemes.  (a) XYZ sequence, (b) ZXY sequence, (c) 3D Hilbert curve, and (d) our height-prioritized 2D Hilbert expansion. Small cubes of varying colors, ranging from red to purple, represent different voxels. The corresponding colored edges indicate the adjacency of voxels that become neighbors after reordering. In each sub-figure, its top half represents the isometric view, and the bottom half represents the top view. Our method maximizes z-axis proximity while also striving to preserve xy-axis proximity.}
    \label{fig:reorder}
    \vspace{-5mm}
\end{figure*}

\subsection{\Name encoder with height-prioritized reordering} 
\label{subsec:mamba}

Semantic occupancy prediction is challenging due to the high dimensionality and density of voxel grids, often involving millions of voxels. Previous methods~\cite{tpvformer, cotr} resort to projection techniques but suffer from information loss. In contrast, Mamba’s linear computational complexity enables direct processing of large voxel features, avoiding the limitations of deformable attention and convolution, such as the need for key point selection or limited receptive fields.

\noindent \textbf{Height-prioritized reordering scheme.} Before feeding 3D voxel features into the Mamba block, it's necessary to reorder them into 1D sequences. A poor reordering strategy can disrupt the intrinsic spatial relationships between adjacent voxels in the 3D space, especially when dealing with a large number of voxel grids, thereby impacting Mamba's performance.
Inspired by the Hilbert curve~\cite{hilbert}, we propose a height-prioritized 2D Hilbert expansion, tailored to the flat spatial structure typical of semantic occupancy prediction tasks, where height information provides valuable categorical clues, as it often correlates with object categories, reveals terrain features, and distinguishes different spatial regions in the scene. 
Specifically, we divide the "xyz" coordinates into the "xy" plane and a "z" dimension. Starting on the xy-plane at z=0, the process extends vertically along the z-axis, creating vertical sequences. These sequences are then ordered and interconnected following the 2D Hilbert curve on the xy plane, forming the height-prioritized 2D Hilbert expansion, as illustrated in Fig.~\ref{fig:reorder}(d). For comparison, Fig.~\ref{fig:reorder}(a)-(c) illustrate the results of reordering according to the 'xyz' sequence, the 'zxy' sequence, and the 3D Hilbert curve, respectively. Our reordering strategy prioritizes z-axis spatial information while maintaining strong spatial proximity in the xy plane. By this way, we can better leverage Mamba's contextual modeling. 

As a result, with the spatial size $W$, $H$, $D$ and 1D data length $L$, the input $\mathbf{V} \in \mathbb{R}^{B \times W \times H \times D \times C}$ and the output $\mathbf{V'} \in \mathbb{R}^{B \times L \times C}$ in the above reordering scheme $\mathcal{R}_{3D \rightarrow 1D}$ and $\mathcal{R}_{1D \rightarrow 3D}$ can be expressed as: 

\vspace{-5mm}
\begin{equation} 
\mathbf{V} = \mathcal{R}_{1D \rightarrow 3D}(\mathbf{V'}) = \mathcal{R}_{1D \rightarrow 3D}(\mathcal{R}_{3D \rightarrow 1D}(\mathbf{V}))
\end{equation}


\noindent \textbf{Hierarchical mamba module.}
To effectively utilize the contextual information across spatial resolutions, we introduce the hierarchical Mamba module, which includes an encoder and decoder. The encoder comprises multiple groups, each with two Mamba blocks and a downsampling operation between groups for multi-scale voxel features. Correspondingly, the Mamba decoder mirrors this structure, utilizing paired Mamba blocks and upsampling operations. The upsampling incorporates skip connection from the corresponding layer in the Mamba encoder, ensuring consistency in feature dimensions across scales. Crucially, the reordering scheme is applied both before and after each Mamba blocks group, maintaining voxel representation during both downsampling and upsampling phases. 

As to a single Mamba block, it consists of LayerNorm(LN), linear layers, 1D convolution, Silu activation, Selective SSM, and residual connections. Given an input $\mathbf{V} \in \mathbb{R}^{B \times L \times C}$, where $B$, $L$ and $C$ denote the batch size, the length of 1D data, and the feature dimension, respectively, the output  $\mathbf{V'} \in \mathbb{R}^{B \times L \times C}$ can be computed as below:

\vspace{-5mm}
\begin{align}
\mathbf{V'_{1}} &= \text{LN}(\mathbf{V}), \label{mamba_block:1} \\
\mathbf{V'_{2}} &= \text{Selective SSM}\left(\text{Silu}(\text{Conv1d}(\text{Linear}(\mathbf{V'_{1}})))\right), \label{mamba_block:2} \\
\mathbf{V'} &= \text{Linear}(\mathbf{V'_{2}} \odot \text{Silu}(\text{Linear}(\mathbf{V'_{1}}))), \label{mamba_block:3}
\end{align}

\noindent where $\odot$ represents the dot product. For convenience, we denote Eq.~\eqref{mamba_block:1}-Eq.~\eqref{mamba_block:3} as $\mathcal{M}$. Then, Given the input $\mathbf{V} \in \mathbb{R}^{B \times W \times H \times D \times C}$ and output $\mathbf{V'} \in \mathbb{R}^{B \times W \times H \times D \times C}$, the computation for a single Mamba block with our reordering scheme $\mathcal{R}_{3D \rightarrow 1D}$ and $\mathcal{R}_{1D \rightarrow 3D}$ can be represented as

\vspace{-1mm}
\begin{equation}
\mathbf{V'} = \mathcal{R}_{1D \rightarrow 3D} \big( \mathcal{M}(\mathcal{R}_{3D \rightarrow 1D}(\mathbf{V})) \big)
\end{equation}

\noindent \textbf{Local context processor.} 
Although the hierarchical Mamba module has improved the network’s ability to process multi-scale global information, handling more local information can further enhance the accuracy of semantic occupancy prediction. To this end, we design a lightweight local context processor. 
It divides the Mamba features $\mathbf{V}_{\mathcal{M}} \in \mathcal{R}^{B \times W \times H \times D \times C}$ along the XY plane into patches $\mathcal{V} = \{\mathcal{V}_{s_i, w_i} | s_i \in S, w_i \in W\}$, using a list of window sizes $W=\{w_i\}$ and a corresponding list of sliding strides $S=\{s_i\}$ to scan the plane. For each item $\mathbf{V}_{p,q} \in \mathcal{V}_{s_i, w_i}$, we have the following equations:  

\vspace{-5mm}
\begin{align}
\mathbf{V}_{p,q} &= 
\mathbf{V}_{\mathcal{M}}[:, p \cdot s_i : p \cdot s_i + w_i, q \cdot s_i : q \cdot s_i + w_i, :, :], 
\nonumber \\
p &= 0, 1,  ... \left\lfloor \frac{W-w_i}{s_i} \right\rfloor,
q = 0, 1, ... \left\lfloor \frac{H-w_i}{s_i} \right\rfloor.
\end{align}

Next, we reorder $\mathbf{V}_{p,q} \in \mathcal{R}^{B \times w_i \times w_i \times D \times C}$ to $\mathcal{R}^{B \times L \times C}$, and then stack $\mathcal{V}_{s_i,w_i}$ to $\mathcal{R}^{(B \times N) \times L \times C}$. Each $\mathcal{V}_{s_i,w_i}$ is processed by the dedicated two-layer Mamba blocks, and the outputs are reshaped to $\mathcal{V'}_{s_i, w_i} = \{\mathbf{V'}_{p,q} \in \mathcal{R}^{B \times w_i \times w_i \times D \times C}\}$. Finally, $\mathcal{V'} = \{\mathcal{V'}_{s_i, w_i}\}$ patches are reassembled into the original spatial size and concatenated along the channel dimension to produce $\mathbf{V'}_{\mathcal{L}} \in \mathcal{R}^{B \times W \times H \times D \times C'}$. A 3D convolution with a kernel size of $1 \times 1 \times 1$ is then applied along the channel dimension to reduce it, resulting in $\mathbf{V}_{\mathcal{L}} \in \mathcal{R}^{B \times W \times H \times D \times C}$.

\vspace{-5mm}
\begin{align}
\mathcal{V'}_{s_i, w_i} &= 
\text{Reshape}(\mathcal{M}(\text{Stack}(\text{Reorder}(\mathcal{V}_{s_i, w_i})))) 
\\
V_\mathcal{L} & = 
\text{Conv3d}(\text{Reassemble\_Patches}(\mathcal{V'})) 
\end{align}


Endowed with the hierarchical Mamba module and local context processor, our \Name utilizes both global and local information from dense scene grids, thereby capturing broad context while preserving local details. This integrated approach enhances spatial understanding and mitigates potential performance degradation with large data volumes, by effectively balancing global awareness with local precision.

\definecolor{barriercolor}{RGB}{255, 0, 0}
\definecolor{bicyclecolor}{RGB}{0, 255, 0}
\definecolor{buscolor}{RGB}{255, 255, 0}
\definecolor{carcolor}{RGB}{0, 0, 255}
\definecolor{construction_vehiclecolor}{RGB}{255, 165, 0}
\definecolor{motorcyclecolor}{RGB}{128, 0, 128}
\definecolor{pedestriancolor}{RGB}{255, 192, 203}
\definecolor{traffic_conecolor}{RGB}{255, 69, 0}
\definecolor{trailercolor}{RGB}{192, 192, 192}
\definecolor{truckcolor}{RGB}{139 ,69 ,19}
\definecolor{driveable_surfacecolor}{RGB}{135 ,206 ,235}
\definecolor{other_flatcolor}{RGB}{160 ,82 ,45}
\definecolor{sidewalkcolor}{RGB}{211 ,211 ,211}
\definecolor{terraincolor}{RGB}{139 ,105 ,20}
\definecolor{manmadecolor}{RGB}{112 ,128 ,144}
\definecolor{vegetationcolor}{RGB}{34 ,139 ,34}

\newcommand{\makecolorcell}[2]{%
    \makecell[b]{\begin{turn}{90}\colorbox{#1}{\hspace{3pt}\rule{0pt}{3pt}} #2\end{turn}}}

\begin{table*}[t]
\centering
\resizebox{\linewidth}{!}{
\setlength{\tabcolsep}{1mm}
\begin{threeparttable}
\begin{tabular}{l|c|cc|cccccccccccccccc}
\toprule
\makecell[b]{Method} & \makecell[b]{Input\\Modality} & 
\makecell[b]{IoU} & \makecell[b]{mIoU} & 
\makecolorcell{barriercolor}{barrier} & 
\makecolorcell{bicyclecolor}{bicycle} & 
\makecolorcell{buscolor}{bus} & 
\makecolorcell{carcolor}{car} & 
\makecolorcell{construction_vehiclecolor}{const. veh.} & 
\makecolorcell{motorcyclecolor}{motorcycle} & 
\makecolorcell{pedestriancolor}{pedestrian} & 
\makecolorcell{traffic_conecolor}{traffic cone} & 
\makecolorcell{trailercolor}{trailer} & 
\makecolorcell{truckcolor}{truck} & 
\makecolorcell{driveable_surfacecolor}{drive surf.} & 
\makecolorcell{other_flatcolor}{other\_flat} & 
\makecolorcell{sidewalkcolor}{sidewalk} & 
\makecolorcell{terraincolor}{terrain} & 
\makecolorcell{manmadecolor}{manmade} & 
\makecolorcell{vegetationcolor}{vegetation} \\

\midrule

MonoScene~\cite{monoscene} & C & 18.4 & 6.9 & 7.1 & 3.9 & 9.3 & 7.2 & 5.6 & 3.0 & 5.9 & 4.4 & 4.9 & 4.2 & 14.9 & 6.3 & 7.9 & 7.4 & 10.0 & 7.6 \\
TPVFormer~\cite{tpvformer} & C & 15.3 & 7.8 & 9.3 & 4.1 & 11.3 & 10.1 & 5.2 & 4.3 & 5.9 & 5.3 & 6.8 & 6.5 & 13.6 & 9.0 & 8.3 & 8.0 & 9.2 & 8.2 \\
SparseOcc~\cite{sparseocc} & C & 21.8 & 14.1 & 16.1 & 9.3 & 15.1 & 18.6 & 7.3 & 9.4 & 11.2 & 9.4 & 7.2 & 13.0 & 31.8 & 21.7 & 20.7 & 18.8 & 6.1 & 10.6 \\
3DSketch~\cite{3dsketch} & C\&D & 25.6 & 10.7 & 12.0 & 5.1 & 10.7 & 12.4 & 6.5 & 4.0 & 5.0 & 6.3 & 8.0 & 7.2 & 21.8 & 14.8 & 13.0 & 11.8 & 12.0 & 21.2 \\
AICNet~\cite{aicnet} & C\&D & 23.8 & 10.6 & 11.5 & 4.0 & 11.8 & 12.3 & 5.1 & 3.8 & 6.2 & 6.0 & 8.2 & 7.5 & 24.1 & 13.0 & 12.8 & 11.5 & 11.6 & 20.2 \\
LMSCNet~\cite{lmscnet} & L & 27.3 & 11.5 & 12.4 & 4.2 & 12.8 & 12.1 & 6.2 & 4.7 & 6.2 & 6.3 & 8.8 & 7.2 & 24.2 & 12.3 & 16.6 & 14.1 & 13.9 & 22.2 \\
JS3C-Net~\cite{js3cnet} & L & 30.2 & 12.5 & 14.2 & 3.4 & 13.6 & 12.0 & 7.2 & 4.3 & 7.3 & 6.8 & 9.2 & 9.1 & 27.9 & 15.3 & 14.9 & 16.2 & 14.0 & 24.9 \\
M-CONet~\cite{openoccupancy} & C\&L & 29.5 & 20.1 & 23.3 & 13.3 & 21.2 & 24.3 & 15.3 & 15.9 & 18.0 & 13.3 & 15.3 & 20.7 & 33.2 & 21.0 & 22.5 & 21.5 & 19.6 & 23.2 \\
Co-Occ~\cite{co-occ} & C\&L & 30.6 & 21.9 & 26.5 & 16.8 & 22.3 & 27.0 & 10.1 & 20.9 & 20.7 & 14.5 & 16.4 & 21.6 & 36.9 & 23.5 & 5.5 & 23.7 & 20.5 & 23.5 \\


\midrule
\textbf{\Name-128 (ours)} & C\&L & \underline{34.7} & \underline{25.2} & \underline{29.1} & \underline{19.1} & \underline{25.5} & \underline{28.5} & \underline{18.1} & \underline{24.7} & \underline{23.4} & \underline{19.8} & \underline{19.3} & \underline{24.5} & \underline{37.0} & \underline{25.4} & \underline{25.4} & \underline{25.4} & \underline{28.1} & \underline{29.9} \\
\textbf{\Name-384 (ours)} & C\&L & \textbf{35.7} & \textbf{26.2} & \textbf{30.2} & \textbf{20.5} & \textbf{26.5} & \textbf{29.5} & \textbf{18.8} & \textbf{26.0} & \textbf{23.7} & \textbf{19.9} & \textbf{20.6} & \textbf{25.4} & \textbf{38.4} & \textbf{26.5} & \textbf{27.0} & \textbf{26.6} & \textbf{28.9} & \textbf{30.5} \\

\bottomrule
\end{tabular}
\end{threeparttable}
}
\caption{Quantitative comparisons on OpenOccupancy validation set with v0.0 annotations. C, D, L denote camera, depth and LiDAR, respectively. \Name-384 means \Name with mamba feature dimension being 384. The best and second-best are in bold and underlined, respectively.}
\label{table_openoccupancy}
\vspace{-3mm}
\end{table*}


\begin{table}[t]
\centering
\fontsize{9pt}{11pt}\selectfont
\begin{threeparttable}
\begin{tabular}{l|c|c} 
\toprule
Method & Input Modality & mIoU \\

\midrule

MonoScene~\cite{monoscene} & C & 11.1 \\
SurroundOcc~\cite{surroundocc} & C & 11.9 \\
OccFormer~\cite{occformer} & C & 12.3 \\
RenderOcc~\cite{renderocc} & C & 12.8 \\
LMSCNet~\cite{lmscnet} & L & 17.0 \\
JS3C-Net~\cite{js3cnet} & L & 23.8 \\
SSC-RS~\cite{SSC-RS} & L & 24.2 \\
Co-Occ~\cite{co-occ} & C\&L & \underline{24.4} \\
M-CONet~\cite{openoccupancy} & C\&L & 20.4  \\

\midrule
\textbf{\Name-128 (ours)} & C\&L & \textbf{24.6}\\

\bottomrule
\end{tabular}
\end{threeparttable}
\caption{Performance on SemanticKITTI test set. The best and second-best are in bold and underlined, respectively.}
\label{table_semantickitti}
\vspace{-3mm}
\end{table}


\begin{table}[t]
\centering
\fontsize{9pt}{11pt}\selectfont
\begin{tabular}{l|c|c} 
\toprule

Method & \makecell[b]{Input\\Modality}  & mIoU  \\

\midrule
SSCNet~\cite{sscnet} & L & 15.2 \\
LMSCNet~\cite{lmscnet} & L & 16.5 \\
MotionSC~\cite{motionsc} & L & 17.6 \\
JS3C-Net~\cite{js3cnet} & L & \underline{22.7} \\

\midrule
\textbf{\Name-128 (ours)} & L & \textbf{23.4} \\

\bottomrule
\end{tabular}
\caption{Comparisons on SemanticPOSS validation set. The best is in bold, and the second-best is underlined.}
\label{table_semanticposs}
\vspace{-3mm}
\end{table}

\subsection{Training objective}
\label{subsec:obj}

Our ultimate training objective is composed of five terms, including the cross-entropy loss $\mathcal{L}_{\text{CE}}$, the lovasz-softmax loss $\mathcal{L}_{\text{iou}}$~\cite{lovasz-softmax}, the geometric and semantic affinity loss $\mathcal{L}_{\text{geo}}$ and $\mathcal{L}_{\text{sem}}$~\cite{monoscene}, and the depth supervision loss $\mathcal{L}_{\text{depth}}$~\cite{bevdepth}. These terms are crucial for optimizing our model's performance in various aspects: $\mathcal{L}_{\text{CE}}$ ensures accurate classification, $\mathcal{L}_{\text{iou}}$ enhances semantic segmentation, $\mathcal{L}_{\text{geo}}$ improves spatial alignment, $\mathcal{L}_{\text{sem}}$ refines semantic understanding, and $\mathcal{L}_{\text{depth}}$ guides depth estimation and spatial relationships. The ultimate training objective is presented as follows:

\vspace{-5mm}
\begin{equation}
\mathcal{L} = \mathcal{L}_{\text{CE}} + \lambda_{1} \mathcal{L}_{\text{iou}} + \lambda_{2} \mathcal{L}_{\text{geo}} + \lambda_{3} \mathcal{L}_{\text{sem}} + \lambda_{4} \mathcal{L}_{\text{depth}},
\end{equation}
where $\lambda_{1} \sim \lambda_{4}$ are the loss coefficients to balance the effect of each loss item on the final performance. They are all empirically set to 1.

\section{Experiments}
\label{sec:experiment}



\subsection{Experimental setup} \label{Experimental Setup}

\noindent \textbf{Benchmarks.} We conduct experiments on OpenOccupancy~\citep{openoccupancy}, SemanticKITTI~\citep{semantickitti} and SemanticPOSS~\cite{semanticposs}. OpenOccupancy is built upon the nuScenes dataset~\citep{nuscenes}, inheriting the data format of nuScenes. It comprises 700 training sequences and 150 validation sequences, with annotations for 17 classes. The occupancy annotations are represented in a 512$\times$512$\times$40 voxel grid, with each voxel sized at 0.2 meters. Notably, each frame in OpenOccupancy has a data range four times larger than the other datasets, imposing a heavier computational burden. In SemanticKITTI, sequences 00-10 (excluding 08), 08, and 11-21 are allocated for training, validation, and testing, respectively. For the occupancy annotations, a 256$\times$256$\times$32 grid is used, with voxels measuring 0.2 meters each. After pre-processing, a total of 19 classes are utilized for training and evaluation. SemanticPOSS, which closely resembles SemanticKITTI, employs sequences 00-05 and 02 as training and validation sets, respectively, with annotations for 11 classes.

\definecolor{barriercolor}{RGB}{255, 0, 0}
\definecolor{bicyclecolor}{RGB}{0, 255, 0}
\definecolor{buscolor}{RGB}{255, 255, 0}
\definecolor{carcolor}{RGB}{0, 0, 255}
\definecolor{construction_vehiclecolor}{RGB}{255, 165, 0}
\definecolor{motorcyclecolor}{RGB}{128, 0, 128}
\definecolor{pedestriancolor}{RGB}{255, 192, 203}
\definecolor{traffic_conecolor}{RGB}{255, 69, 0}
\definecolor{trailercolor}{RGB}{192, 192, 192}
\definecolor{truckcolor}{RGB}{139 ,69 ,19}
\definecolor{driveable_surfacecolor}{RGB}{135 ,206 ,235}
\definecolor{other_flatcolor}{RGB}{160 ,82 ,45}
\definecolor{sidewalkcolor}{RGB}{211 ,211 ,211}
\definecolor{terraincolor}{RGB}{139 ,105 ,20}
\definecolor{manmadecolor}{RGB}{112 ,128 ,144}
\definecolor{vegetationcolor}{RGB}{34 ,139 ,34}

\newcommand{\legenditem}[2]{\raisebox{0.5ex}{\colorbox{#1}{\rule{0pt}{0.4ex}\rule{0.4ex}{0pt}}} \hspace{0.1em}{#2}\hspace{0.25em}}

\begin{figure*}[t]
    \centering
    \includegraphics[width=\linewidth]{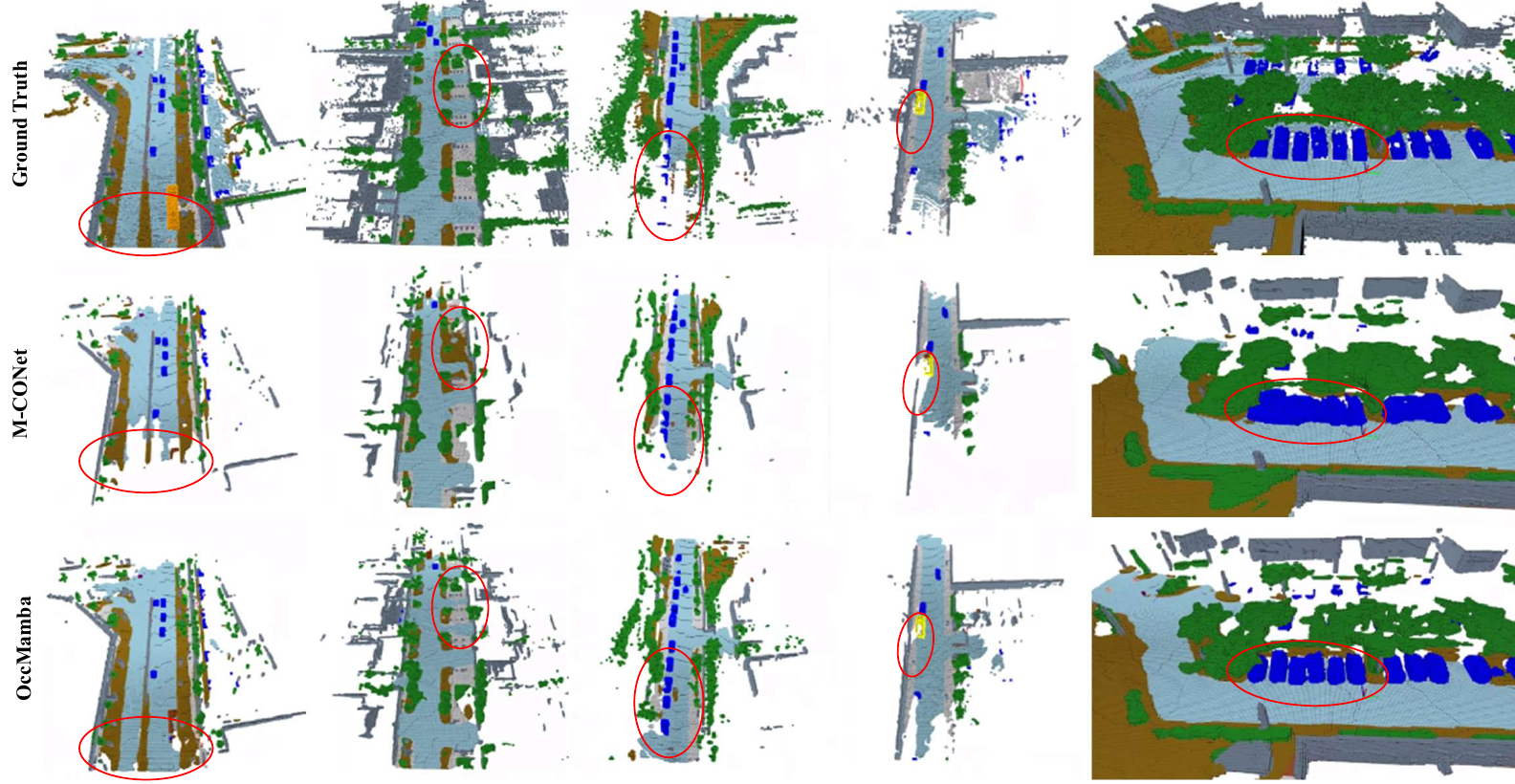}

    \parbox{\textwidth}{
    \centering
    \legenditem{barriercolor}{Barrier} 
    \legenditem{bicyclecolor}{Bicycle} 
    \legenditem{buscolor}{Bus} 
    \legenditem{carcolor}{Car} 
    \legenditem{construction_vehiclecolor}{Construction Vehicle} 
    \legenditem{motorcyclecolor}{Motorcycle} 
    \legenditem{pedestriancolor}{Pedestrian} 
    \legenditem{traffic_conecolor}{Traffic Cone}
    }
    
    
    \parbox{\textwidth}{
    \centering
    \legenditem{trailercolor}{Trailer} 
    \legenditem{truckcolor}{Truck} 
    \legenditem{driveable_surfacecolor}{Driveable Surface} 
    \legenditem{other_flatcolor}{Other Flat} 
    \legenditem{sidewalkcolor}{Sidewalk} 
    \legenditem{terraincolor}{Terrain} 
    \legenditem{manmadecolor}{Manmade} 
    \legenditem{vegetationcolor}{Vegetation}
    }
    
    \caption{Visual comparison between OccMamba and M-CONet. From top to bottom: ground-truth, predictions of M-CONet and OccMamba. Our OccMamba makes preciser predictions than M-CONet especially in regions highlighted by red ellipses.}
    \label{fig:openoccupancy_results}
\end{figure*}

\noindent \textbf{Evaluation metrics.} We adopt the official evaluation metrics, \emph{i.e.}, Intersection-over-Union (IoU) and mean Intersection-over-Union (mIoU).

\noindent \textbf{Implementation details.} We use ResNet-50~\citep{resnet} as the image backbone. Both the Mamba Encoder and Mamba Decoder in the hierarchical mamba module consist of four groups and each group contains two Mamba blocks. The local context processor utilizes window sizes of [3, 5, 7], with two Mamba blocks implemented. For the OpenOccupancy dataset, we maintain identical setting to M-CONet~\citep{openoccupancy}: for each frame, we utilize six surround view camera images as input, coupled with a fusion of ten frames of LiDAR points spanning the range of [-51.2m, 51.2m] along the X and Y axes, and [-2.0m, 6.0m] along the Z axis. Within the Mamba block, experiments are conducted with mamba feature dimension set to 128 and 384, respectively. On the SemanticKITTI dataset, our inputs consist of forward-facing stereo camera images alongside a single-frame LiDAR points, with their spatial extents defined as [0m, -25.6m, -2m, 51.2m, 25.6m, 4.4m]. We set the mamba feature dimension to 128 and employ Test Time Augmentation (TTA) to boost the performance. The used augmentations include the flipping of the X and Y axes, and four augmented inputs (including the original input) are sent to the model. Finally, for the SemanticPOSS, we exclusively leverage single-frame LiDAR points like SemanticKITTI.


\subsection{Experimental results} \label{Quantitative results}

\noindent \textbf{OpenOccupancy.} As shown in Table~\ref{table_openoccupancy}, we compare \Name on OpenOccupancy with v0.0 annotation to previous methods which use the settings of camera-only~\citep{monoscene, tpvformer, 3dsketch, aicnet}, LiDAR-only~\citep{lmscnet, js3cnet}, and multi-modal fusion~\citep{openoccupancy, co-occ}. 
Our \Name achieves the best performance. Particularly, \Name-384 outperforms Co-Occ~\citep{co-occ} by 5.1\% in IoU and 4.3\% in mIoU, respectively. Additionally, \Name-128 also surpasses Co-Occ by 4.1\% in IoU and 3.3\% in mIoU. These results underscore the efficacy of our approach in handling large-scale scenarios. 

\noindent \textbf{SemanticKITTI \& SemanticPOSS.} As evident from Table~\ref{table_semantickitti}, our \Name-128 still outperforms the second-best method, \emph{i.e.}, Co-Occ~\citep{co-occ}, by 0.2\% mIoU in SemanticKITTI test set. Note that we do not compare our \Name with SCPNet~\cite{scpnet} and Occfiner~\cite{occfiner} as they employ tricks such as label rectification, knowledge distillation or multi-frame concatenation. Strong performance is also observed in SemanticPOSS, as shown in Table~\ref{table_semanticposs}. These results sufficiently demonstrate the superiority of our approach. The detailed classwise performance is put in the supplementary material.


\noindent \textbf{Visual comparisons.}
From Fig.~\ref{fig:openoccupancy_results}, our \Name-384 produces more accurate predictions than M-CONet~\cite{openoccupancy} on OpenOccupancy validation set. Our \Name not only provides a more detailed representation of object shapes and semantics, but also performs preciser predictions on occluded objects and surfaces.

\begin{table}[t]
\centering
\fontsize{9pt}{11pt}\selectfont
\begin{tabular}{l|cc}
\toprule
Reordering Schemes & mIoU \\

\midrule


XYZ sequence & 24.5  \\
ZXY sequence & 24.7 \\
3D Hilbert & 24.8 \\
\textbf{Height-prioritized 2D Hilbert expansion} & \textbf{25.2} \\

\bottomrule
\end{tabular}
\caption{Ablation study on different reordering schemes.}
\label{table_ablation_study_reorder}
\vspace{-1mm}
\end{table}

\begin{table}[t]
\centering
\resizebox{\linewidth}{!}{
\begin{tabular}{l|c|c|c|c}
\toprule
Method & CNN & \makecell[b]{Hierarchical\\Mamba\\Module} & \makecell[b]{Local\\Context\\Processor}  & mIoU  \\

\midrule
M-CONet~\cite{openoccupancy} & \checkmark &  &  & 20.1 \\
\textbf{\Name-128} &  & \checkmark &  & 25.0 \\
\textbf{\Name-128} &  & \checkmark & \checkmark  & 25.2 \\
\textbf{\Name-384} &  & \checkmark & \checkmark  & \textbf{26.2} \\

\bottomrule
\end{tabular}
}
\caption{Ablation study on OccMamba Encoder.}
\label{table_ablation_occmamba_encoder}
\vspace{-4mm}
\end{table}

\subsection{Ablation study} \label{Ablation study}
The reported results are on the OpenOccupancy validation set with v0.0 annotations unless otherwise specified.

\noindent \textbf{Reordering schemes.} To accelerate the training process, we set the mamba feature dimension to 128 and then compare our method with three alternative methods: ordering by XYZ sequence, ordering by ZXY sequence, and ordering using the 3D Hilbert curve. As shown in Table~\ref{table_ablation_study_reorder}, our method achieves the highest mIoU of 25.2\%. Additionally, it is apparent that reordering schemes that prioritize the z-axis outperform the other methods. This could be attributed to the fact that in semantic occupancy prediction, the overall shape of the scenes typically manifests as a flattened cuboid. Therefore, reordering schemes that prioritize the z-axis allow Mamba to better capture spatial proximity information, leading to improved prediction performance. 


\noindent \textbf{\Name encoder.} 
Consistent with previous settings, we conduct ablation studies on mamba feature dimension and two modules within our \Name encoder. As shown in Table~\ref{table_ablation_occmamba_encoder}, by replacing the CNN-based occupancy encoder from M-CONet~\cite{openoccupancy} with our hierarchical Mamba module using a mamba feature dimension of 128, the mIoU improves from 20.1\% to 25.0\%. The incorporation of our local context processor further enhances the performance, achieving 25.2\% mIoU. When the Mamba feature dimension is increased from 128 to 384, the mIoU improves further, reaching 26.2\%.

\noindent \textbf{Generalization.} We apply our hierarchical Mamba module to the MonoScene\cite{monoscene} framework. Specifically, we replace the 3D voxel decoder in this framework with our implementation, using a mamba feature dimension of 128, and train on 25\% of the SemanticKITTI training set with the training strategy unchanged. As shown in Table~\ref{table_ablation_study_backbones}, our \Name encoder can improve the performance of MonoScene from 11.1\% mIoU to 11.9\% mIoU on SemanticKITTI val set.



\noindent \textbf{Memory usage and inference time.} \label{Memory Usage and Inference Time}
In Table~\ref{table_memory_and_time}, we present a comparative analysis of memory usage and inference time between \Name and M-CONet during both training and inference phases. The networks are trained on 8 A40 GPUs and perform inference on a single RTX 4090 GPU.When the mamba feature dimension is set to 384, \Name shows similar memory usage to M-CONet during training but reduces inference memory usage by about 24\%. When the mamba feature dimension is lowered to 128, the memory savings are more pronounced, with a reduction of approximately 38\% during training and 44\% during inference compared to M-CONet. Additionally, \Name-128 significantly improves inference speed, taking about 35\% of the time required by M-CONet, which highlights its efficiency in resource-constrained scenarios. In terms of performance, \Name-384 achieves the highest mIoU, while \Name-128 configurations also surpass M-CONet in performance, providing a balance between efficiency and accuracy.

\begin{table}[t]
\centering
\resizebox{\linewidth}{!}{
\begin{tabular}{l|c|c|c|c}
\toprule
Method & CNN & \makecell[b]{Hierarchical\\Mamba\\Module} & \makecell[b]{Local\\Context\\Processor}  & mIoU  \\

\midrule
MonoScene~\cite{monoscene} & \checkmark &  &  & 11.1  \\
MonoScene~\cite{monoscene} & & \checkmark &  & 11.7 \\
MonoScene~\cite{monoscene} & & \checkmark & \checkmark & \textbf{11.9} \\

\bottomrule
\end{tabular}
}
\caption{Ablation study on MonoScene framework.}
\label{table_ablation_study_backbones}
\vspace{-1mm}
\end{table}

\begin{table}[t]
\centering
\resizebox{\linewidth}{!}{
\begin{tabular}{l|cc|c|c}
\toprule
\multirow{2}{*}{Method} & \multicolumn{2}{c|}{Memory Usage(GiB)} & \multirow{2}{*}{\begin{tabular}[c]{@{}c@{}}Inference\\ Time(ms)\end{tabular}} & \multirow{2}{*}{mIoU} \\

 & \multicolumn{1}{c|}{Training} & Inference & & \\
 
\midrule
M-CONet~\cite{openoccupancy} & \multicolumn{1}{c|}{37.3} & 12.1 & 416.0 & 20.1 \\

\midrule
\textbf{\Name-384} & \multicolumn{1}{c|}{37.7} & 9.2 & 448.3 & \textbf{26.2} \\
\textbf{\Name-128} & \multicolumn{1}{c|}{\textbf{23.1}} & \textbf{6.8} & \textbf{269.3} & 25.2 \\

\bottomrule
\end{tabular}
}
\caption{Comparison of memory usage and inference time. } 
\label{table_memory_and_time}
\vspace{-3mm}
\end{table}
\section{Conclusion}
\label{sec:conclusion}

In this paper, we present the first Mamba-based network, termed \Name, for semantic occupancy prediction. To facilitate the processing of Mamba blocks and maximally retain the 3D spatial relationship, we design a novel reordering scheme and OccMamba encoder. 
Endowed with these designs, our \Name is capable of directly and efficiently processing large volumes of dense scene grids, surpassing the previous state-of-the-art algorithms on three prevalent occupancy prediction benchmarks. As an innovative approach in this field, our \Name provides an effective means for handling large-scale voxels directly, and we believe it will inspire new advancements.
{
    \small
    \bibliographystyle{ieeenat_fullname}
    \bibliography{main}
}

\clearpage
\setcounter{page}{1}
\maketitlesupplementary

\section{Appendix section}
\label{sec:appendix_section}

\subsection{More training details}
\label{More training details}

In our experiment, we utilize the AdamW optimizer with a base learning rate of $5\mathrm{e}{-4}$ and a weight decay of 0.01 to ensure effective optimization while maintaining regularization. As to the image backbone ResNet-50, which is pre-trained by torchvision, we scale its learning rate by a factor of 0.1. The learning rate schedule follows a Cosine Annealing policy, combined with a linear warmup over the first 500 iterations, starting from one-third of the base learning rate, and gradually decreasing to a minimum ratio of $1\mathrm{e}{-3}$. For training, we set the number of epochs to 20. This experimental setup reflects our focus on balancing optimization efficiency, model stability, and rigorous evaluation to achieve reliable and reproducible results.

\subsection{More reordering schemes}
\label{More reordering schemes}

In addition to the Hilbert curve, other space-filling curves, such as the Z-order, are also widely used. Therefore, we conduct comparative experiments, following the procedures outlined in Sec.~\ref{Experimental Setup} and Sec.~\ref{Ablation study} on the OpenOccupancy validation set with v0.0 annotations to evaluate the performance of the Z-order. As presented in Table~\ref{table_more_reorder}, it is evident that our \Name-128 with height-prioritized 2D Hilbert expansion outperforms the Z-curve variant in semantic occupancy prediction. In theory, the Hilbert curve effectively preserves the spatial proximity when mapped to a 1D sequences due to its recursive, space-filling path. In contrast, the Z-order curve employs a simple interleaving of bits, which makes it more likely that adjacent points are separated by greater distances in the 1D sequences. Consequently, the Hilbert curve generally offers superior locality preservation in multiple dimensions.

\begin{table}[htbp]
\centering
\fontsize{9pt}{11pt}\selectfont
\begin{tabular}{l|cc}
\toprule
Reordering Schemes & mIoU \\

\midrule

3D Z-order & 24.6 \\
3D Hilbert & 24.8 \\
Height-prioritized 2D Z-order expansion & 25.0 \\
\textbf{Height-prioritized 2D Hilbert expansion} & \textbf{25.2} \\

\bottomrule
\end{tabular}
\vspace{-2mm}
\caption{Performance on more reordering schemes.}
\label{table_more_reorder}
\vspace{-5mm}
\end{table}

\subsection{More ablations on local context processor (LCP)}
\label{More ablations on local context processor}

\noindent \textbf{Metric Specificity.} OpenOccupancy's mIoU does not use techniques like visual masks, causing ambiguous evaluation of occluded regions. In Table.~\ref{table:suppl_lcp_ablations}, LCP improves IoU (denser occupancy) and RayIoU (from SparseOcc~\cite{sparseocc}, surface accuracy, excluding occlusions) by \textbf{1.0\%} and \textbf{0.6\%} in v0.0 labels, respectively, validating its effectiveness in refining geometric coherence (Fig.~\ref{fig:suppl_lcp_visual}(a,b)).

\noindent \textbf{Label Quality Impact.} The old OpenOccupancy labels (v0.0), derived from static LiDAR, suffer from incomplete annotations for dynamic objects (Fig.~\ref{fig:suppl_lcp_visual}(c)) and occluded areas LiDAR never seen. By using new labels (v0.1), 
our LCP improves mIoU by \textbf{0.4\%}(Table.~\ref{table:suppl_lcp_ablations}),
demonstrating more gains as label noise reduces.

\newcommand{\localwidth}{0.155\textwidth}

\begin{figure}[htbp]
    \centering
    \setlength{\tabcolsep}{0.5pt}
    \begin{tabular}{ccc}
        \includegraphics[width=\localwidth]{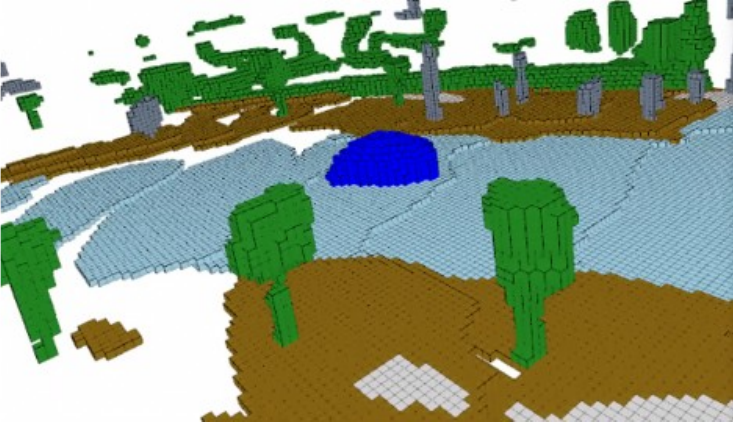} &
        \includegraphics[width=\localwidth]{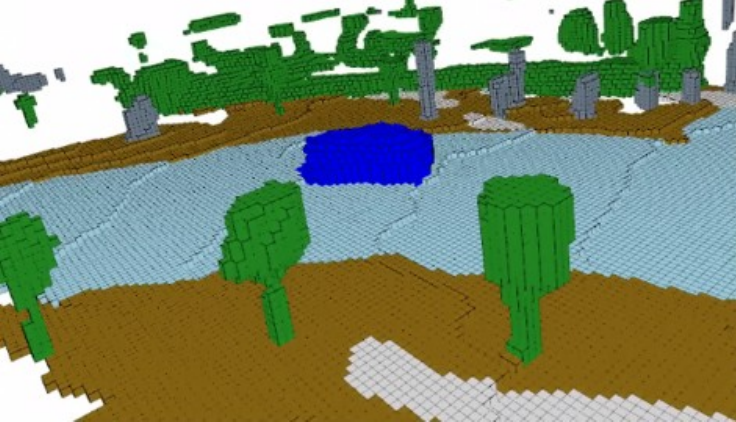} &
        \includegraphics[width=\localwidth]{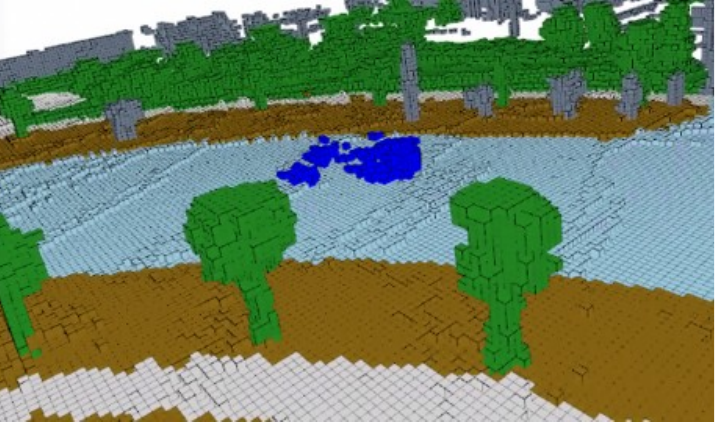} \\
        \parbox[c]{\localwidth}{\centering (a) w/o LCP} &
        \parbox[c]{\localwidth}{\centering (b) w LCP} &
        \parbox[c]{\localwidth}{\centering (c) label} \\
    \end{tabular}
    \vspace{-3mm}
    \caption{Reconstruction results of distant occupancy (about 40m).}
    \label{fig:suppl_lcp_visual}
    \vspace{-7mm}
\end{figure}
\begin{table}[htbp]
\centering
\begin{tabular}{|c|c|c|c|c|}
\hline
Method  & Label & IoU  & mIoU & RayIoU@0.2m \\ \hline
w/o LCP & v0.0  & 33.7 & 25.0 & 24.2        \\ \hline
w LCP   & v0.0  & 34.7 & 25.2 & 24.7        \\ \hline
w/o LCP & v0.1  & 34.2 & 25.8 & 26.5        \\ \hline
w LCP   & v0.1  & 34.9 & 26.2 & 27.0        \\ \hline
\end{tabular}
\vspace{-2mm}
\caption{More results of OccMamba-128.}
\label{table:suppl_lcp_ablations}
\vspace{-7mm}
\end{table}

\subsection{Ablation on training loss}
\label{Training objectives analysis}

We conduct an ablation study on the training objectives mentioned in Sec.~\ref{subsec:obj}. Specifically, we carry out experiments on the OpenOccupancy dataset, following the procedures outlined in Sec.~\ref{Experimental Setup}. To facilitate training, we use only 20\% of the training set, with the model configured as \Name-128. 
The results, as shown in Table~\ref{table_ablation_study_loss}, indicate that all the training objectives contribute significantly to the ultimate performance. In particular, the inclusion of $\mathcal{L}_{\text{iou}}$ and $\mathcal{L}_{\text{CE}}$ yield significant performance enhancements, as evidenced by the increase in mIoU, highlighting their critical role in our \Name.

\begin{table}[htbp]
\centering
\fontsize{9pt}{11pt}\selectfont
\begin{tabular}{c|c|c|c|c|c}
\toprule
$\mathcal{L}_{\text{CE}}$ & $\mathcal{L}_{\text{iou}}$ & 
$\mathcal{L}_{\text{depth}}$ &
$\mathcal{L}_{\text{geo}}$ & $\mathcal{L}_{\text{sem}}$ & mIoU \\

\midrule
 & & & & \checkmark & 19.2 \\
 & & & \checkmark & \checkmark & 19.5 \\
 & & \checkmark & \checkmark & \checkmark & 19.9 \\
 & \checkmark & \checkmark & \checkmark & \checkmark & 21.7 \\
\checkmark & \checkmark & \checkmark & \checkmark & \checkmark & 22.9 \\

\bottomrule
\end{tabular}
\vspace{-2mm}
\caption{Ablation study on the effect of each training loss.}
\label{table_ablation_study_loss}
\vspace{-7mm}
\end{table}

\subsection{More experimental results}
\label{More experimental results}

Due to space constraints, we put the detailed class-wise performance on the SemanticKITTI dataset in this section. As shown in Table~\ref{table_semantickitti_full}, our OccMamba achieves state-of-the-art results on SemanticKITTI test set.

\definecolor{emptycolor}{RGB}{0, 0, 0}
\definecolor{carcolor}{RGB}{245, 150, 100}
\definecolor{bicyclecolor}{RGB}{245, 230, 100}
\definecolor{motorcyclecolor}{RGB}{150, 60, 30}
\definecolor{truckcolor}{RGB}{180, 30, 80}
\definecolor{other-vehiclecolor}{RGB}{255, 0, 0}
\definecolor{personcolor}{RGB}{30, 30, 255}
\definecolor{bicyclistcolor}{RGB}{200, 40, 255}
\definecolor{motorcyclistcolor}{RGB}{90, 30, 150}
\definecolor{roadcolor}{RGB}{255, 0, 255}
\definecolor{parkingcolor}{RGB}{255, 150, 255}
\definecolor{sidewalkcolor}{RGB}{75, 0, 75}
\definecolor{other-groundcolor}{RGB}{75, 0, 175}
\definecolor{buildingcolor}{RGB}{0, 200, 255}
\definecolor{fencecolor}{RGB}{50, 120, 255}
\definecolor{vegetationcolor}{RGB}{0, 175, 0}
\definecolor{trunkcolor}{RGB}{0, 60, 135}
\definecolor{terraincolor}{RGB}{80, 240, 150}
\definecolor{polecolor}{RGB}{150, 240, 255}
\definecolor{traffic-signcolor}{RGB}{0, 0, 255}

\begin{table*}[htbp]
\centering
\fontsize{9pt}{11pt}\selectfont
\setlength{\tabcolsep}{0.5mm}
\begin{threeparttable}
\begin{tabular}{l|c|c|ccccccccccccccccccc} 
\toprule
Method & \makecell[b]{Input\\Modality} & \makecell[b]{mIoU} & \makecell[b]{\begin{turn}{90}\colorbox{roadcolor}{\hspace{3pt}\rule{0pt}{3pt}} road\end{turn}} & \makecell[b]{\begin{turn}{90}\colorbox{sidewalkcolor}{\hspace{3pt}\rule{0pt}{3pt}} sidewalk\end{turn}} & \makecell[b]{\begin{turn}{90}\colorbox{parkingcolor}{\hspace{3pt}\rule{0pt}{3pt}} parking\end{turn}} & \makecell[b]{\begin{turn}{90}\colorbox{other-groundcolor}{\hspace{3pt}\rule{0pt}{3pt}} other ground\end{turn}} & \makecell[b]{\begin{turn}{90}\colorbox{buildingcolor}{\hspace{3pt}\rule{0pt}{3pt}} building\end{turn}} & \makecell[b]{\begin{turn}{90}\colorbox{carcolor}{\hspace{3pt}\rule{0pt}{3pt}} car\end{turn}} & \makecell[b]{\begin{turn}{90}\colorbox{truckcolor}{\hspace{3pt}\rule{0pt}{3pt}} truck\end{turn}} & \makecell[b]{\begin{turn}{90}\colorbox{bicyclecolor}{\hspace{3pt}\rule{0pt}{3pt}} bicycle\end{turn}} & \makecell[b]{\begin{turn}{90}\colorbox{motorcyclecolor}{\hspace{3pt}\rule{0pt}{3pt}} motorcycle\end{turn}} & \makecell[b]{\begin{turn}{90}\colorbox{other-vehiclecolor}{\hspace{3pt}\rule{0pt}{3pt}} other vehicle\end{turn}} & \makecell[b]{\begin{turn}{90}\colorbox{vegetationcolor}{\hspace{3pt}\rule{0pt}{3pt}} vegetation\end{turn}} & \makecell[b]{\begin{turn}{90}\colorbox{trunkcolor}{\hspace{3pt}\rule{0pt}{3pt}} trunk\end{turn}} & \makecell[b]{\begin{turn}{90}\colorbox{terraincolor}{\hspace{3pt}\rule{0pt}{3pt}} terrain\end{turn}} & \makecell[b]{\begin{turn}{90}\colorbox{personcolor}{\hspace{3pt}\rule{0pt}{3pt}} person\end{turn}} & \makecell[b]{\begin{turn}{90}\colorbox{bicyclistcolor}{\hspace{3pt}\rule{0pt}{3pt}} bicyclist\end{turn}} & \makecell[b]{\begin{turn}{90}\colorbox{motorcyclistcolor}{\hspace{3pt}\rule{0pt}{3pt}} motorcyclist\end{turn}} & \makecell[b]{\begin{turn}{90}\colorbox{fencecolor}{\hspace{3pt}\rule{0pt}{3pt}} fence\end{turn}} & \makecell[b]{\begin{turn}{90}\colorbox{polecolor}{\hspace{3pt}\rule{0pt}{3pt}} pole\end{turn}} & \makecell[b]{\begin{turn}{90}\colorbox{traffic-signcolor}{\hspace{3pt}\rule{0pt}{3pt}} traffic sign\end{turn}} \\

\midrule

MonoScene~\cite{monoscene} & C & 11.1 & 54.7 & 27.1 & 24.8 & 5.7 & 14.4 & 18.8 & 3.3 & 0.5 & 0.7 & 4.4 & 14.9 & 2.4 & 19.5 & 1.0 & 1.4 & 0.4 & 11.1 & 3.3 & 2.1 \\
SurroundOcc~\cite{surroundocc} & C & 11.9 & 56.9 & 28.3 & 30.2 & 6.8 & 15.2 & 20.6 & 1.4 & 1.6 & 1.2 & 4.4 & 14.9 & 3.4 & 19.3 & 1.4 & 2.0 & 0.1 & 11.3 & 3.9 & 2.4 \\
OccFormer~\cite{occformer} & C & 12.3 & 55.9 & 30.3 & 31.5 & 6.5 & 15.7 & 21.6 & 1.2 & 1.5 & 1.7 & 3.2 & 16.8 & 3.9 & 21.3 & 2.2 & 1.1 & 0.2 & 11.9 & 3.8 & 3.7 \\
RenderOcc~\cite{renderocc} & C & 12.8 & 57.2 & 28.4 & 16.1 & 0.9 & 18.2 & 24.9 & 6.0 & 0.4 & 0.3 & 3.7 & 26.2 & 4.9 & 3.6 & 1.9 & 3.1 & 0.0 & 9.1 & 6.2 & 3.4 \\
LMSCNet~\cite{lmscnet} & L & 17.0 & 64.0 & 33.1 & 24.9 & 3.2 & 38.7 & 29.5 & 2.5 & 0.0  & 0.0 & 0.1 & 40.5 & 19.0 & 30.8 & 0.0 & 0.0 & 0.0 & 20.5 & 15.7 & 0.5 \\
JS3C-Net~\cite{js3cnet} & L & 23.8 & 64.0 & 39.0 & 34.2 & \underline{14.7} & 39.4 & 33.2 & \underline{7.2} & \textbf{14.0} & \textbf{8.1} & \textbf{12.2} & 43.5 & 19.3 & 39.8 & \textbf{7.9} & \textbf{5.2} & 0.0 & 30.1 & 17.9 & 15.1 \\
SSC-RS~\cite{SSC-RS} & L & 24.2 & \textbf{73.1} & \textbf{44.4} & \underline{38.6} & \textbf{17.4} & \textbf{44.6} & \underline{36.4} & 5.3 & \underline{10.1} & 5.1 & \underline{11.2} & \underline{44.1} & 26.0 & \textbf{41.9} & \underline{4.7} & 2.4 & \textbf{0.9} & 30.8 & 15.0 & 7.2 \\
Co-Occ~\cite{co-occ} & C\&L & \underline{24.4} & \underline{72.0} & \underline{43.5} & \textbf{42.5} & 10.2 & 35.1 & \textbf{40.0} & 6.4 & 4.4 & 3.3 & 8.8 & 41.2 & \textbf{30.8} & \underline{40.8} & 1.6 & \underline{3.3} & 0.4 & \textbf{32.7} & \underline{26.6} & \underline{20.7} \\
M-CONet~\cite{openoccupancy} & C\&L & 20.4 & 60.6 & 36.1 & 29.0 & 13.0 & 38.4 & 33.8 & 4.7 & 3.0 & 2.2 & 5.9 & 41.5 & 20.5 & 35.1 & 0.8 & 2.3 & \underline{0.6} & 26.0 & 18.7 & 15.7 \\

\midrule
\textbf{OccMamba-128 (ours)} & C\&L & \textbf{24.6} & 68.7 & 41.0 & 35.9 & 9.1 & \underline{40.8} & 34.8 & \textbf{8.8} & 8.8 & \underline{6.5} & 8.9 & \textbf{44.9} & \underline{28.7} & 40.6 & 4.2 & 2.6 & \underline{0.6} & \underline{32.0} & \textbf{27.0} & \textbf{23.3} \\

\bottomrule
\end{tabular}
\end{threeparttable}
\caption{Performance on SemanticKITTI test set. The best and second-best are in bold and underlined, respectively.}
\label{table_semantickitti_full}
\end{table*}

\end{document}